\documentclass[nohyperref]{article}

\usepackage{microtype}
\usepackage{graphicx}
\usepackage{subfigure}
\usepackage{booktabs} % for professional tables
\usepackage{pgfplots}

\usepackage{hyperref}

\usepackage[accepted]{icml2022}

\usepackage{amsmath}
\usepackage{amssymb}
\usepackage{mathtools}
\usepackage{amsthm}
\usepackage{tabularx}

\usepackage[capitalize,noabbrev]{cleveref}

\theoremstyle{plain}

\theoremstyle{definition}

\theoremstyle{remark}

\usepackage[textsize=tiny]{todonotes}

\icmltitlerunning{LAVA: Language Audio Vision Alignment for Data-Efficient Video Pre-Training}

\begin{document}

\twocolumn[
\icmltitle{LAVA: Language Audio Vision Alignment for Data-Efficient Video Pre-Training}

% It is OKAY to include author information, even for blind
% submissions: the style file will automatically remove it for you
% unless you've provided the [accepted] option to the icml2022
% package.

% List of affiliations: The first argument should be a (short)
% identifier you will use later to specify author affiliations
% Academic affiliations should list Department, University, City, Region, Country
% Industry affiliations should list Company, City, Region, Country

% You can specify symbols, otherwise they are numbered in order.
% Ideally, you should not use this facility. Affiliations will be numbered
% in order of appearance and this is the preferred way.
\icmlsetsymbol{equal}{*}

\begin{icmlauthorlist}
\icmlauthor{Sumanth Gurram}{berkeley}
\icmlauthor{Andy Fang}{berkeley}
\icmlauthor{David Chan}{berkeley}
\icmlauthor{John Canny}{berkeley}
%\icmlauthor{}{sch}
%\icmlauthor{}{sch}
\end{icmlauthorlist}

\icmlaffiliation{berkeley}{University of California, Berkeley, USA}

\icmlcorrespondingauthor{Sumanth Gurram}{sumanthgurram@berkeley.edu}
\icmlcorrespondingauthor{David Chan}{davidchan@berkeley.edu}

% You may provide any keywords that you
% find helpful for describing your paper; these are used to populate
% the "keywords" metadata in the PDF but will not be shown in the document
\icmlkeywords{Machine Learning, ICML}

\vskip 0.3in
]

% this must go after the closing bracket ] following \twocolumn[ ...

% This command actually creates the footnote in the first column
% listing the affiliations and the copyright notice.
% The command takes one argument, which is text to display at the start of the footnote.
% The \icmlEqualContribution command is standard text for equal contribution.
% Remove it (just {}) if you do not need this facility.

\printAffiliationsAndNotice{}  % leave blank if no need to mention equal contribution
%\printAffiliationsAndNotice{\icmlEqualContribution} % otherwise use the standard text.

\begin{abstract}
Generating representations of video data is of key importance in advancing the field of machine perception. Most current techniques rely on hand-annotated data, which can be difficult to work with, expensive to generate, and hard to scale. In this work, we propose a novel learning approach based on contrastive learning, LAVA, which is capable of learning joint language, audio, and video representations in a self-supervised manner. We pre-train LAVA on the Kinetics 700 dataset using transformer encoders to learn representations for each modality. We then demonstrate that LAVA performs competitively with the current state-of-the-art self-supervised and weakly-supervised pre-training techniques on UCF-101 and HMDB-51 video action recognition while using a fraction of the unlabeled data.
\end{abstract}

%%%%%%%%% INTRODUCTION
\section{Introduction}
\label{section:introduction}

Supervised learning has generally driven the progress video representation learning, however, labeling datasets is both time-consuming and expensive, making it especially hard to leverage large amounts of data using supervised learning. Moreover, while attention-based architectures such as \citet{dosovitskiy2020image} \citet{arnab2021vivit, bertasius2021space} have started to outperform CNNs on key benchmarks, they often require much larger amounts of training data than CNNs. 

Self-supervised methods have emerged to answer the challenge; as powerful pre-training strategies for vision tasks, they can scale to larger training datasets without being constrained by labeling needs. One common approach is to use data augmentation to learn representational invariants for vision \citet{qian2021spatiotemporal, jing2018self}. Instead of relying on hand-designed augmentations for the visual modality, another approach is to exploit the multi-modal nature of video and learn audio-visual correspondence, as seen with contrastive methods such as those in \citet{morgado2020audio,patrick2021space,patrick2020multi,morgado2020learning,korbar2018cooperative}. Similarly, other methods use text metadata from videos to learn joint visual-text representations in a self/weakly-supervised manner \citet{stroud2020learning,patrick2020support,li2020learning,sun2019learning,miech2020end,sun2019videobert}. 

\begin{figure}
    \includegraphics[width=\columnwidth]{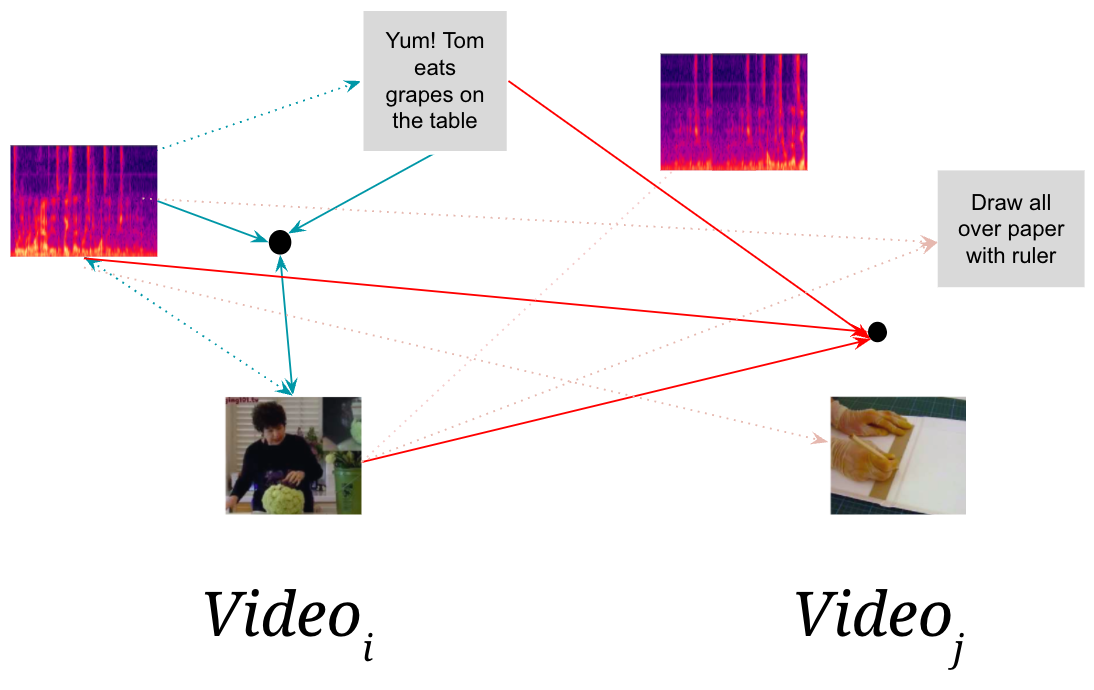}
    \caption{For a given sample, the video, audio, and text data are all encoded into embeddings. LAVA's pre-training objective involves contrasting video embeddings from one sample to audio and text embeddings from different samples (dotted red lines) while aligning embeddings from the same sample (dotted green lines). Additionally, LAVA will calculate a centroid from audio, video, and text embeddings for each sample and contrast (solid red lines) or align (solid green lines) embeddings to these centroids accordingly.}
    \label{fig:lava_objective}
\end{figure}

\begin{figure*}
    \centering
    \includegraphics[height=6.1cm]{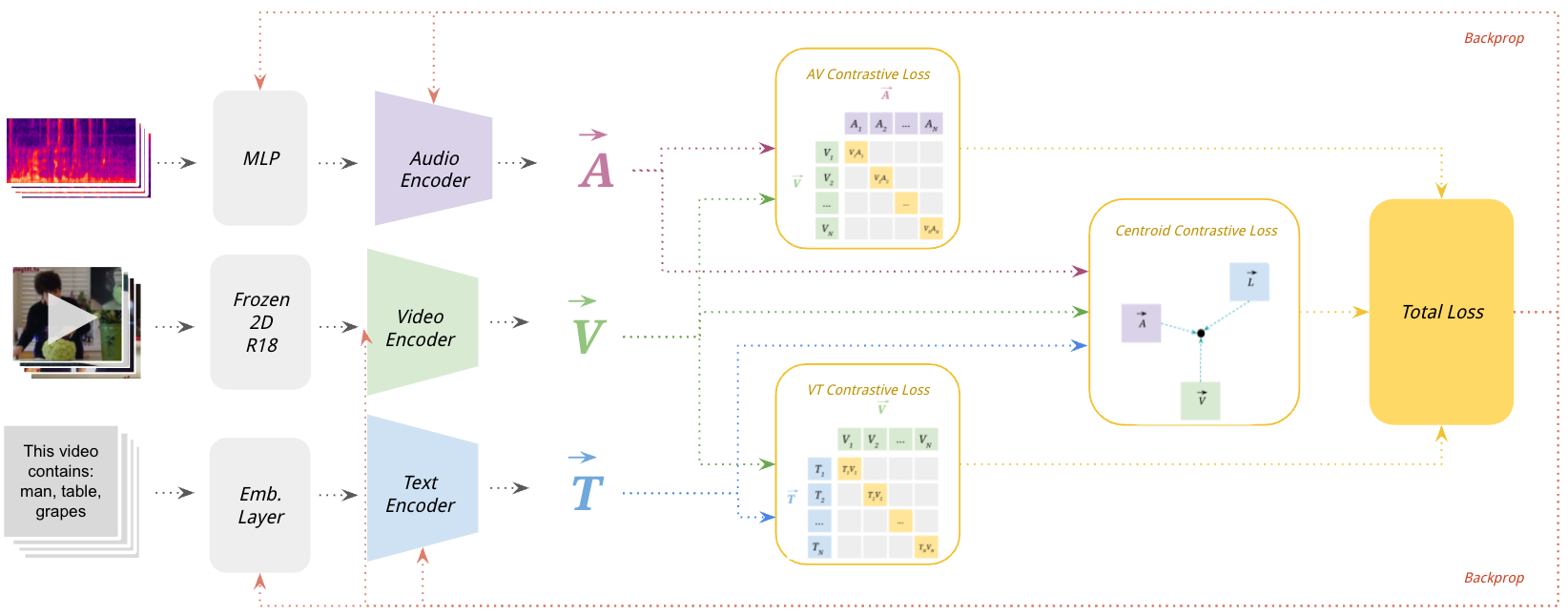}
    \caption{LAVA's pre-training architecture includes modality-specific feature extraction, attention-based encoding, cross-modal contrastive losses and centroid contrastive loss. Implementation details are in Section \ref{section:experiments}.}
    \label{fig:lava_architecture}
\end{figure*}

A less common but even more label-efficient strategy is to learn audio, visual and text representations together. While methods such as \citet{akbari2021vatt,alayrac2020self,chen2021multimodal} pre-train on these three modalities, they do so leveraging HowTo100M, which has a massive 15 years of unlabeled video data. Moreover, excepting \citet{akbari2021vatt}, most of the above techniques use highly modality-specific encoders (e.g. only CNNs for vision), instead of exploring more generic, attention-based backbones for all modalities. 

These observations are the main motivation for LAVA, which introduces a more data and label-efficient method for pre-training transformer encoders on audio, visual, and text modalities for video data. To achieve this, our proposed method uses novel cross-modal and centroid-based contrastive objectives seen in Figure \ref{fig:lava_objective}. We evaluate pre-trained LAVA on UCF-101 and HMDB-51.

%%%%%%%%% APPROACH
\section{Pre-Training Approach}
\label{section:approach}

\paragraph{Overview} 
Given a set of $n$ unlabelled videos $X$, each video $x_i \in X$  is decomposed into different modalities $a_i$, $v_i$ and $t_i$, which are audio, visual and text features, respectively. Details regarding modality-specific extraction are in Section \ref{section:experiments}. Then, LAVA's three encoders $f_a$, $f_v$, and $f_t$ produce output embeddings given their respective modality features $m_i$ as inputs. LAVA then uses projection functions to project these embeddings to various multi-modal latent spaces (e.g. audio-video, video-text and audio-video-text spaces) in some $\mathbb{R}^d$. Let us denote $z_{m}$ as the embedding for modality $m$. We define projection function $g_{m,m'}$ to project $z_{m}$ and $z_{m'}$, to a multi-modal latent space  $m,m'$. In these multi-modal latent spaces, we apply a contrastive framework to jointly compare embeddings using a similarity function $s$, such that $\forall i, {m'} \neq {m}$ we have high $s(g_{m,m'}(z_{m,i}), g_{m,m'}(z_{m',i})$ and $\forall i \neq j, {m'} \neq {m}$ we have low $s(f_m(m_i), f_{m'}({m'}_j))$.

\paragraph{Cross-Modal Contrastive Loss} We use the function $$s(\cdot,\cdot) =  exp(g_{m,m'}(z_{m,i})^T g_{m,m'}(z_{m',j})/\tau)$$ as the similarity function, where $\tau$ denotes temperature, and the noise contrastive estimation (NCE) \citet{pmlr-v9-gutmann10a} as our contrastive loss, where positive pairs are embeddings from the same instance and negatives are embeddings from difference instances. This objective is intended to make LAVA a dictionary, effectively mapping input features $a_i, v_i, t_i$ to unified representations $z_i$ in a multi-modal latent space. The NCE loss is formulated below (for simplicity the projection functions are omitted):
\begin{equation}
\label{eq:nce}
  \resizebox{\columnwidth}{!}{
        $NCE(z_{m}, z_{m'}) = -log(\frac{\sum_{i=0}^{N} exp(z_{m,i}^T z_{m',i}/\tau)}{\sum_{i=0}^{N} exp(z_{m,i}^T z_{m',i}/\tau) + \sum_{i=0}^{N} \sum_{j\neq{i}}^{N} exp(z_{m,i}^T z_{m',j}/\tau)})$
    }
\end{equation}

Following \citet{miech2020end}, we use NCE for audio-video and video-text pairs: 
\begin{equation}
\label{eq:av_loss}
    L_{AV}(z_{a}, z_{v}) = NCE(g_{av}(z_{a}), g_{av}(z_{v})) 
\end{equation}
\vspace{-1.5em}
\begin{equation}
\label{eq:vt_loss}
    L_{VT}(z_{v}, z_{t}) = NCE(g_{vt}(z_{v}), g_{vt}(z_{t}))
\end{equation}

\paragraph{Centroid Contrastive Loss} LAVA also enforces audio-video-text correspondence via a novel centroid contrastive loss. Following \citet{chen2021multimodal}, for a given instance $x_i$, we calculate a centroid $c_i$ by averaging projected LAVA embeddings:  $c_i = (g_{avt}(z_{a,i}) + g_{avt}(z_{v,i}) + g_{avt}(z_{t,i})) / 3$. $g_{avt}$ projects modal embeddings to a joint, tri-modal latent space. However, unlike \citet{chen2021multimodal} we do not k-means cluster the centroids via k-means clustering and align embeddings to their centroid's cluster assignment. Thus, we avoid potential detriments of grouping representations into a fixed K clusters across all batches and the additional training time needed for clustering. Instead, we instead directly optimize for alignment between embeddings and their centroid via the following loss:
\begin{equation}
\label{eq:centroid_loss}
    L_{AVT}(c, z_a, z_v, z_t) = \sum_{m \in {a,v,t}} NCE(g_{avt}(z_{m}), c)
\end{equation}
Combining all losses, we define the total loss for LAVA's pre-training:
\begin{equation}
\label{eq:total_loss}
    L_{LAVA} = L_{AV} + L_{VT} + L_{AVT}
\end{equation}

%%%%%%%%% EXPERIMENTS
\section{Experiments}
\label{section:experiments}

\begin{table*}[t]
  \centering
  \footnotesize
%   \resizebox{\columnwidth}{!}{
  \begin{tabularx}{\linewidth}{Xlllll}
    \toprule
    Method & Dataset (years) & GPU Hours & Mod. & UCF & HMDB \\
    \midrule
    RotNet3D \citet{jing2018self} & K600 (0.1) & - & V & 47.7 & 24.8 \\
    CBT \citet{sun2019learning} & K600 (0.1) & 1536 & VT & 54.0 & 29.5 \\
    MemDPC \citet{han2020memory} & K600 (0.1) & - & VF & 54.1 & 30.5 \\
    AVSF \citet{xiao2020audiovisual} & K400 (0.1) & - & AV & 54.1 & 30.5 \\
    CoCLR \citet{han2020self} & K400 (0.1) & - & VF & 77.4 & 44.1 \\
    STiCA* \citet{patrick2021space}  & K400 (0.1) & 2930 & AV & 77.0 & 48.2 \\
    CPD \citet{li2020learning}  & IG300 (0.1) & - & VT & 83.7 & 54.7 \\
    CVRL \citet{qian2021spatiotemporal}  & K600 (0.1) & - & V & 90.8 & 59.7 \\
    \hline
    LAVA* (video only) & K700 (0.1) & 408 & AVT & 81.3 & 50.1 \\
    LAVA* (audio+video) & K700 (0.1) & 408 & AVT & 84.3 & - \\
    \hline
    MIL-NCE \citet{miech2020end} & HT (15) & 4608 & VT & 83.4 & 54.8 \\
    ELo \citet{piergiovanni2020evolving} & Y8M & 4608 & V & - & 64.5 \\
    VATT* \citet{akbari2021vatt} & HT (15) & 18432 & AVT & 89.6 & 65.2 \\
    BraVe \citet{recasens2021broaden} & AS (1) & - & AV & 93.6 & 70.8 \\
    MMV$^\dagger$ \citet{alayrac2020self} & AS+HT (16) & 2304 & AVT & 95.2 & 75.0 \\
    WTS \citet{stroud2020learning} & WTS70M (22) & 9216 & VT & 95.8 & 77.7 \\
    \bottomrule
  \end{tabularx}
%   }
  \caption{* denotes that the vision backbone has a transformer encoder. $^\dagger$ denotes that the vision backbone is fine-tuned downstream, rather than being frozen for linear evaluation. The duration of each dataset is also measured in years.}
  \vspace{0.5em}
  \label{tab:main_results}
\end{table*}

\paragraph{Pre-training} We use the training set of Kinetics-700: 480k videos of which 300k videos have usable audio and web-crawled titles. To the best of our knowledge, LAVA is the first method to extract this text data for Kinetics-700 and we intend to release this data for others to use. Video clips are [16 x 224 x 224 x 3] at 10 fps, with random h-flip and space-crop augmentations. Audio features are [80, 256] log mel-spectrograms augmented with Gaussian noise. Text sequences have a maximum length of 128 from a 48k vocab size using BPE tokenization. Most video transformers use ImageNet-pre-trained ViT \citet{arnab2021vivit,bertasius2021space}, pre-train on massive video datasets like HowTo100M as in \citet{akbari2021vatt} or use convolutions followed transformers as in \citet{patrick2021space}. We follow the latter, except we freeze a ImageNet-pre-trained ResNet-18 backbone and use it to extract frame-wise patch features for each clip. Audio features are encoded by an MLP. Text tokens are mapped to embeddings. As seen in Figure \ref{fig:lava_architecture}, audio, video and text features are then encoded by transformer encoders $f_a$, $f_v$, and $f_t$, each with 4 layers and a hidden size of 1024. Missing modalities in a batch are masked out during loss calculation. We use a batch size of 32, 0.07 temperature, $1e^{-5}$ learning rate, Adam optimizer, and a cosine-learning schedule. Pre-training is done for 25 epochs on a single Titan X GPU.

\paragraph{UCF-101 Downstream} The UCF-101 dataset has 13k videos across 101 action categories. We train a linear classifier on top of the frozen LAVA video encoder using a 1e-4 learning rate and a batch size of 32. Also, we compare the downstream performance of the classifier when using the frozen video encoder vs. frozen audio and video encoders. Logits are averaged across multiple clips per test set video. Top-1 accuracy is averaged across all 3 splits.

\paragraph{HMDB-51 Downstream} The HMDB-51 dataset has 7k videos across 51 action categories. We follow the UCF-101 evaluation procedure, except we do not evaluate with audio+video mode as most HMDB-51 videos have no audio. Otherwise, our procedure follows UCF-101.

\section{Results}

\begin{table}
  \centering
  \footnotesize
    % \resizebox{\columnwidth}{!}{
  \begin{tabularx}{\linewidth}{Xlll}
    \toprule
    Pre-train & Downstream & UCF-101 & HMDB-51 \\
    \midrule
    AV & Video Only & 68.6 & 41.5\\
    AV & Audio+Video & 72.4 & -\\
    AV+VT & Video Only & 74.91 & 49.4\\
    AV+VT & Audio+Video & 79.36 & -\\
    AV+VT+AVT & Video Only & 81.1 & 51.8\\
    AV+VT+AVT & Audio+Video & 84.2 & -\\
    \bottomrule
  \end{tabularx}
%   }
  \caption{All results are split-1 top-1 accuracy. Pre-train denotes which pre-training losses are used.}
  \label{tab:modality_ablation}
\end{table}

\paragraph{Downstream Tasks} Table \ref{tab:main_results} compares LAVA linear evaluation performance on UCF-101 and HMDB-51 to various self-supervised benchmarks. LAVA outperforms all methods trained on datasets comparable to Kinetics-700 size besides \citet{li2020learning,qian2021spatiotemporal}. This can be attributed to LAVA's increased data efficiency as it makes use of audio, video, and text for each sample, whereas these benchmarks rely on at most two modalities. Notably, LAVA significantly outperforms \citet{sun2019learning}, which omits audio and uses ASR-extracted captions as text, whereas LAVA is pre-trained on audio and video titles. LAVA also slightly outperforms \citet{patrick2021space}, which has a R(2+1)-18+transformer video encoder and ResNet audio encoder, but does not pre-train on text. LAVA performs competitively with \citet{li2020learning}, which also uses video titles as text, and is outperformed by \citet{qian2021spatiotemporal}, which uses the video modality using extensive spatio-temporal augmentations and a large batch size of 1024. However, both methods \citet{li2020learning,qian2021spatiotemporal} use ResNet3D-50 as their backbone, which is more performant when trained from scratch than transformer encoders trained with the same amount of data \citet{bertasius2021space}. See Section \ref{ssec:future_work} for plans to directly compare for to these methods.

 Interestingly, \citet{miech2020end}, despite having been pre-trained on over 150X samples than Kinetics-700, only slightly outperforms LAVA on UCF-101 and HMDB-51. We believe this is because \citet{miech2020end} uses caption-based text and does not use audio, whereas LAVA is pre-trained on audio and titles-based text. \citet{akbari2021vatt, alayrac2020self} outperform LAVA using audio, video and caption-based text, while \citet{stroud2020learning} uses video and titles-based text. While this performance gap can likely be attributed to LAVA's reliance on less than 1\% of the data used in these benchmarks, it does seem using audio and titles-based text can increase data efficiency as LAVA's performance rivals that of \citet{miech2020end}.

Given that LAVA uses Kinetics-700 video titles as text, it can be argued that this is a form of weak supervision as per \citet{li2020learning, stroud2020learning}. We found that while all HMDB-51 classes are fully covered by Kinetics-700 text, the following UCF-101 classes are not covered: IceDancing, PizzaTossing, PommelHorse, SkiJet, StillRings. The 3-split average top-1 accuracy for these classes is 79.9\%, which is comparable to LAVA's overall 81.3\% for UCF-101, indicating that LAVA's representations transfer well to unseen classes downstream.

Since LAVA is a multi-modal model, we use our novel evaluation strategy to quantify downstream performance improvements when using embeddings from audio and video modalities. Table \ref{tab:main_results} shows that audio-video LAVA outperforms video-only by 3\%, indicating that audio embeddings may provide additional information for action recognition. See Section \ref{ssec:future_work} for more plans in this direction.

Lastly, pre-training LAVA takes 408 GPU hours, which is over 70\% faster than all other benchmarks documented in Table \ref{tab:main_results}. Notably, LAVA also completes all pre-training using just 1 GPU, whereas all other benchmarks use few as 4 and as many as 256 accelerators. This combination of 70\% fewer accelerators and 70\% fewer GPU hours means LAVA is significantly more efficient in terms of computational cost, in addition to label and data-efficiency.

\begin{figure}
\centering
\resizebox{\columnwidth}{!}{
\begin{tikzpicture}
\begin{axis}[
    title={},
    xlabel={\# Pre-training Epochs},
    ylabel={UCF-101 Top-1 Accuracy},
    xmin=0, xmax=30,
    ymin=70, ymax=90,
    xtick={0,5,10,15,20,25,30},
    ytick={70, 75, 80, 85, 90},
    legend pos=north west,
    ymajorgrids=true,
    grid style=dashed,
]

\addplot[
    color=blue,
    mark=square,
    ]
    coordinates {
    (1,73.08)(10,80.04)(25,81.27)
    };
   \addlegendentry{Video Only}
   
\addplot[
    color=red,
    mark=triangle,
    ]
    coordinates {
    (1,75.98)(10,82.31)(25,84.33)
    };
    \addlegendentry{Audio+Video}
\end{axis}
\end{tikzpicture}
}
\vspace{-2em}
\caption{UCF-101 accuracy as a function of pre-training duration.} 
\label{fig:duration_ablation}
\end{figure}
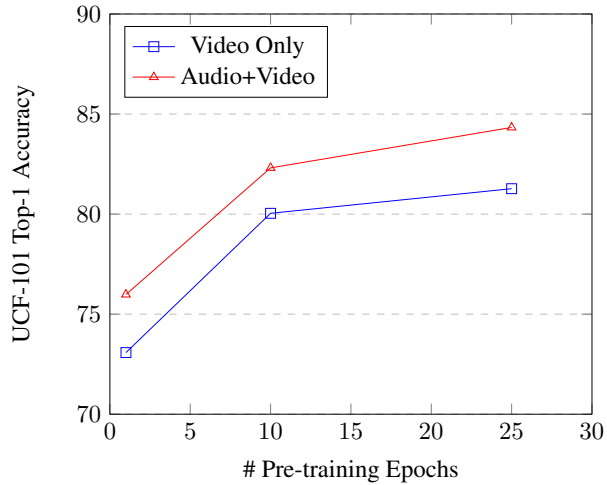

\paragraph{Ablations} As shown in Table \ref{fig:duration_ablation}, we ablate pre-training duration by evaluating LAVA at 1, 10, and 25 epochs of pre-training; video-only increases from 73.1\% to 80.0\% to 81.3\%, while audio+video increases from 76.0\% to 82.3\% to 84.3\%. Additionally, we ablate pre-training losses as seen in Table \ref{tab:modality_ablation}. As expected, LAVA pre-trained without text and centroid contrastive loss performs the worst. LAVA pre-trained with $L_{AV}$ and $L_{VT}$, but no centroid loss performs worse than LAVA pre-trained with the centroid loss. This indicates that the text modality and the centroid loss, both of which increase data efficiency by making greater use of the same number of samples, significantly improve downstream video action recognition. The performance boost from using audio+video embeddings during linear evaluation is also seen in Table \ref{tab:modality_ablation}.

%%%%%%%%% CONCLUSION
\section{Conclusion}
\label{section:conclusion}
\label{ssec:future_work}

We present LAVA, a novel self-supervised pre-training method for learning audio, visual, and text representations from unlabeled video data using transformers. By using multiple modalities and introducing novel cross-modal and centroid contrastive objectives, LAVA increases data efficiency while performing competitively with self and weakly supervised benchmarks. As LAVA's pre-training effectively combines multi-modal data and generic transformer encoders, we believe it can scale both in terms of the amount on unlabelled data and the number of modalities in the data. 

\paragraph{Future Work} To more directly compare LAVA performance to \citet{li2020learning,qian2021spatiotemporal}, we plan to pre-train the transformer vision encoder on ImageNet first or use the ResNet3D-50 encoder from scratch, as well as larger batch sizes for contrastive pre-training and more extensive video augmentation. We also plan to pre-train LAVA on larger datasets with non-title-based text (e.g. HowTo100M) to compare more directly with \citet{akbari2021vatt, alayrac2020self}. Since contrastive pre-training with audio may dilute information in the video embeddings, we plan to compare downstream performance between video embeddings from VT-pretrained LAVA and those of current LAVA. Additionally, we plan to ablate how different coefficients between various LAVA losses will affect downstream performance. Lastly, we intend to explore pre-training on more/different modalities and new downstream tasks (e.g. retrieval, captioning, video/audio generation) using LAVA.

\clearpage

\bibliography{lava}
\bibliographystyle{icml2022}

\end{document}